\pdfoutput=1

\documentclass[11pt]{article}

\usepackage{acl}

\usepackage{times}
\usepackage{latexsym}

\usepackage[T1]{fontenc}

\usepackage[utf8]{inputenc}

\usepackage{microtype}

\usepackage{inconsolata}

\usepackage{graphicx}
\usepackage{multirow}
\usepackage{bm}
\usepackage{mathtools}
\usepackage{amsfonts}
\usepackage{pgfplots}
\usepackage{subfigure}
\usepackage{soul}
\usepackage{color}
\usepackage{booktabs}

\title{LANDeRMT: Detecting and Routing Language-Aware Neurons for Selectively Finetuning LLMs to Machine Translation}

\author{Shaolin Zhu\textsuperscript{1}\footnotemark[2], Leiyu Pan\textsuperscript{1}\footnotemark[2], Bo Li\textsuperscript{2,3}, Deyi Xiong\textsuperscript{1}\footnotemark[1]\\
\textsuperscript{1}College of Intelligence and Computing, Tianjin University, Tianjin, China\\
\textsuperscript{2}School of Software, Tsinghua University, Beijing, China\\
\textsuperscript{3}Baidu APP Technology and Platform R\&D Department, Baidu Inc, Beijing, China\\
\texttt{\{zhushaolin, lypan, dyxiong\}@tju.edu.cn}\\
\texttt{\{li-b19\}@mails.tsinghua.edu.cn}} 

\begin{document}
\maketitle

\renewcommand{\thefootnote}{\fnsymbol{footnote}}
\footnotetext[2]{Equal contribution.}
\footnotetext[1]{Corresponding author.}

\renewcommand{\thefootnote}{\arabic{footnote}}

\begin{abstract}
Recent advancements in large language models (LLMs) have shown promising results in multilingual translation even with limited bilingual supervision.
The major challenges are catastrophic forgetting and parameter interference\footnote{Regarding catastrophic forgetting and parameter interference, we are specifically addressing issues between languages rather than those between machine translation tasks and other tasks in this paper.} for finetuning LLMs when provided parallel training data.
To address these challenges, we propose LANDeRMT, a \textbf{L}anguage-\textbf{A}ware \textbf{N}euron \textbf{De}tecting and \textbf{R}outing framework that selectively finetunes LLMs to \textbf{M}achine \textbf{T}ranslation with diverse translation training data. 
In LANDeRMT, we evaluate the awareness of neurons to MT tasks and categorize them into language-general and language-specific neurons. 
This categorization enables selective parameter updates during finetuning, mitigating parameter interference and catastrophic forgetting issues.
For the detected neurons, we further propose a conditional awareness-based routing mechanism to dynamically adjust language-general and language-specific capacity within LLMs, guided by translation signals.
Experimental results demonstrate that the proposed LANDeRMT is very effective in learning translation knowledge, significantly improving translation quality over various strong baselines for multiple language pairs.

\end{abstract}

\section{Introduction}
\label{Introduction}
Conventional neural machine translation (NMT) usually requires a huge amount of parallel training data \cite{DBLP:journals/corr/abs-2207-04672,DBLP:journals/jmlr/FedusZS22,zhu2023unsupervised,zhu2024mining}.
In contrast, multilingual LLMs, e.g., BLOOM \cite{DBLP:journals/corr/abs-2211-05100}, LLaMA2 \cite{DBLP:journals/corr/abs-2307-09288}, in spite of being trained with mainly monolingual data, require only a few examples to demonstrate remarkable prowess in multilingual translation via in-context learning (ICL) \cite{DBLP:journals/corr/abs-2305-04118,DBLP:journals/corr/abs-2305-01181}. 
However, such LLM-based MT exhibits a major drawback that the quality of yielded translations is highly sensitive to the provided examples in ICL \cite{DBLP:conf/acl/VilarFCLRF23} and outputs might suffer from overgeneration \cite{DBLP:conf/eamt/BawdenY23}.

To address these issues, existing studies attempt to use various finetuning methods, such as adapter-based method \cite{DBLP:conf/emnlp/AlvesGAPRSCM23}, instruction-based tuning method \cite{DBLP:journals/corr/abs-2305-15083}.
However, these approaches primarily focus on balancing between the original LLMs and new finetuning translation data 
They use only incremental data to acquire new knowledge without considering catastrophic forgetting of knowledge  originally captured by LLMs \cite{DBLP:conf/aaai/LiuYW21,DBLP:conf/acl/Shao022,DBLP:conf/acl/HuangLMYL23}.
Many studies have shown that catastrophic forgetting indeed exists across languages as LLMs are fine-tuned on one language pair and then used to translate another language on which LLMs are not fine-tuned \cite{li-etal-2023-mmnmt,zhu-etal-2024-towards-robust-context}.
Additionally, as LLMs are usually generally developed for multiple tasks (i.e., sharing parameters across different tasks), finetuning LLMs for MT task may cause parameter interference for other tasks \cite{DBLP:journals/corr/abs-2308-08747}.
In Section \ref{Main Results}, we find that full-parameter finetuning of LLMs cannot always improve translation quality on all language pairs.
Therefore, is it possible to design a new finetuning method for LLMs, which can mitigate the issues of catastrophic forgetting and parameter interference during the finetuning process of LLMs to multilingual machine translation?

In multilingual NMT, previous efforts evaluate the importance of model neurons to each
language pair and only tune language-specific neurons for the current language pair during training \cite{DBLP:conf/acl/Xie0G020}.
Recent studies on LLM unveils that many neurons in the feed-forward networks (FFN) are only activated for specific tasks and become ``dead'' for irrelevant tasks \cite{DBLP:journals/corr/abs-2309-04827,DBLP:journals/corr/abs-2304-14997}.

Inspired by these studies, we propose LANDeRMT, a language-aware neuron detecting and routing framework for selectively finetuning LLMs to MT, which aims to mitigate the issues of catastrophic forgetting and parameter interference.
First, we evaluate the MT awareness of each neuron in FFNs.
For neurons that are related to multilingual MT tasks, we further evaluate the relevance of each neuron to each language pair. 
According to their MT/language ``awareness/relevance'', we divide neurons into the unactivated neurons, language-general neurons and language-specific neurons.
After that, we finetune LLMs on multilingual parallel training data.
During finetuning, only the parameters of language-general neurons and language-specific neurons for the current language pair are tuned.
This selective finetuning process can alleviate the parameter interference issue.

As language-general and language-specific capacity matters for MT \cite{DBLP:conf/iclr/ZhangBSF21,DBLP:conf/acl/KoishekenovBN23}, we propose a conditional awareness routing mechanism to dynamically schedule language-general and language-specific capacity across sub-layers in LLMs under the guidance of translation signals.
In doing so, we can alleviate the catastrophic forgetting issue and facilitate LLMs to be adapted to MT.

The main contributions of this work are summarized as follows:

\begin{itemize}
\item We propose LANDeRMT that aims at mitigating the catastrophic forgetting and parameter interference issues for efficiently finetuning LLMs to MT.
\item To well schedule language-general and language-specific capacity across sub-layers in LLMs, we propose a conditional awareness-based routing mechanism.
\item Experiments on ten language pairs show that our model achieves the state-of-the-art results compared to previous strong baselines and demonstrate the robustness of the proposed model in various settings.
\end{itemize}

\section{Related Work}
\label{Related Work}

LLMs, with a few examples provided via in-context learning, have demonstrated impressive capabilities in machine translation without requiring explicit supervision from parallel training data \cite{DBLP:conf/eamt/MoslemHKW23,DBLP:journals/corr/abs-2302-07856,DBLP:journals/corr/abs-2305-03573,DBLP:conf/wmt/HanESGN22}. 
However, LLMs with ICL for MT suffer from the sensitiveness to the provided examples \cite{DBLP:conf/acl/VilarFCLRF23} and yielded translations might be overgenerated \cite{DBLP:conf/eamt/BawdenY23}.

Another line of research on LLMs, known as domain-adaptive pretraining, focuses
on finetuning LLMs to downstream tasks \cite{DBLP:journals/corr/abs-2309-09530,DBLP:journals/corr/abs-2310-05492}. 
Although these approaches have demonstrated efficacy in adapting various LLMs and result in enhanced performance on downstream tasks \cite{DBLP:journals/corr/abs-2303-17564,DBLP:journals/corr/abs-2308-04014,DBLP:journals/corr/abs-2401-02415,zhu-xiong-2023-tjunlp}, they rarely apply to  multilingual generation tasks, e.g., multilingual MT.

In order to efficiently adapt LLMs to MT, recent years have witnessed efforts on finetuning LLMs for MT \cite{DBLP:conf/acl/VilarFCLRF23,DBLP:conf/emnlp/AlvesGAPRSCM23}.
\citet{DBLP:conf/emnlp/AlvesGAPRSCM23} show that adapter-based finetuning with LoRA \cite{DBLP:conf/iclr/HuSWALWWC22} matches the performance of traditional finetuning while reducing the number of training parameters by a factor of 50.
\citet{DBLP:journals/corr/abs-2305-15083} investigate the multilingual generalization when finetuning LLMs.
However, they do not explicitly overcome catastrophic forgetting and parameter interference issues.
To address these issues, our work starts with analyzing the neurons within the model, and finetunes LLMs by distinguishing neurons.

\begin{figure*}[t]
\centering
\centerline{\includegraphics[scale=0.52]{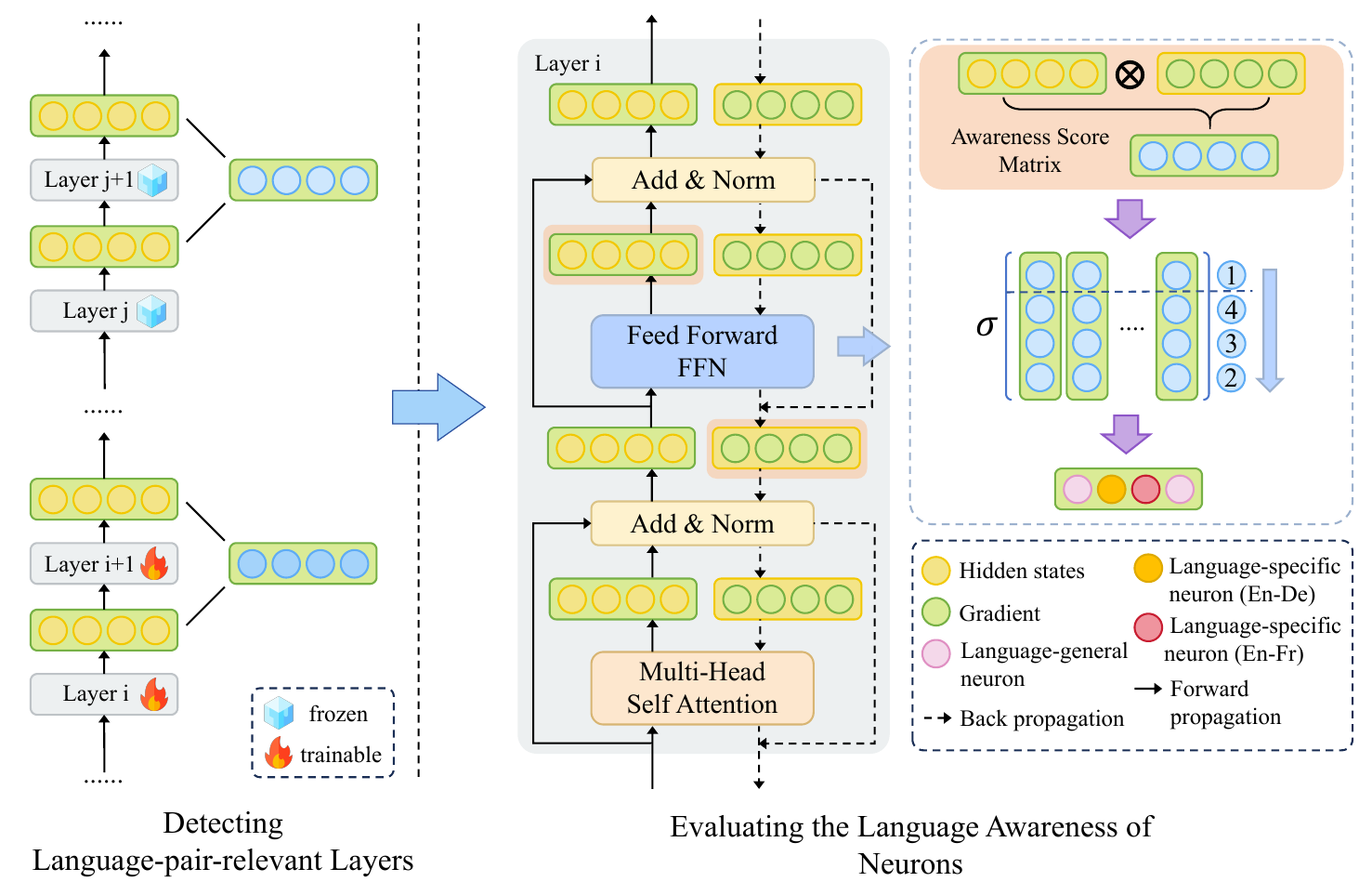}}
\caption{Illustration of the proposed LANDeRMT.
}
\label{fig1}
\end{figure*}

Research interests in understanding the inner workings of LLMs and NMT models have been growing recently \cite{DBLP:journals/corr/abs-2207-13243,bills2023language,DBLP:journals/corr/abs-2310-01870}.
\citet{DBLP:journals/corr/abs-2309-04827} focus on neurons inside FFNs and find that the network is sparse and represents many discrete features in LLMs. 
They find many of the alive neurons are reserved for discrete features and act as token and n-gram detectors for different languages.
In addition, previous NMT efforts evaluate the importance of NMT neurons in each
language pair and only finetune language-specific neurons for the current language pair participate in training for conventional multilingual NMT \cite{DBLP:conf/acl/Xie0G020,DBLP:conf/conll/PatelCA22}.
Partially motivated by these studies, we propose a language-aware neuron detecting and routing framework for selectively finetuning LLMs to MT. 
In our method, we use awareness-based evaluation of neurons in LLMs and divide the neurons into language-general and language-specific neurons.
We only update the parameters of language-general neurons and the corresponding language-specific neurons for the current language pair during training to overcome catastrophic forgetting and parameter interference to enhance the multilingual translation ability of LLMs.

\section{Methodology}
\label{Methodology}

The proposed LANDeRMT is illustrated in Figure \ref{fig1}.
We first propose a method to analyse which FFN layers of LLMs have strong relevance to source-target language pair.
This allows us to exclusively concentrate on layers that are related to the MT task, hence reducing the distraction from unrelated parameters.
Then, we employ Taylor Expansion (TE) \cite{DBLP:conf/acl/Xie0G020} to evaluate the strength of awareness (relevance) of neurons at those layers to the given language pairs of the MT task.
Finally, we only route and finetune the detected language-aware neurons for the MT task.
This can ensure that we only need to update a small number of relevant parameters of LLMs for MT.

\subsection{Detecting Language-Pair-Relevant Layers}
\label{Special Layers Evaluation}

We introduce a representation analysis (RA) method to detect language-pair-relevant layers, which is based on the difference in activations between FFN layers. 
RA aims to measure the changes in the response of each FFN layer to the input source sentence during the LLM forward propagation process that ``translates'' the source sentence into the target sentence, so as to identify FFN alignment layers that are highly ``activated'' for the source-target language pair. 
For each consecutive pair of layers $i$ and $i+1$ within the LLM, we compute the activation difference $D$, to estimate the degree of change in information representation between these two layers. 
The estimation is computed as follows:

\begin{equation}
D_i = \left| \frac{1}{N} \sum_{n=1}^{N} \textbf{A}_{i,n} - \frac{1}{N} \sum_{n=1}^{N} \textbf{A}_{i+1,n} \right|
\end{equation}
where $\textbf{A}_{i,n}$ and $\textbf{A}_{i+1,n}$ represent the activation values at the $i$-th and $i+1$-th layers during the $n$-th forward propagation, and $N$ is the total number of forward propagations. 
In this manner, $D_i$ captures the extent of change in activation values between adjacent layers along the depth of the model when the input source language is translated into the target language.
The most significant changes in layer representations indicate the most critical layers that are related to the source-target language pair. 
Therefore, our layer selection criterion focuses on identifying those layers with the top-$k$ $D$ values as follows:

\begin{equation}
L_{\text{LPR}} = \arg \max_{top_{k}} \{D_1, D_2, ..., D_k\}
\end{equation}
where $L_{\text{LPR}}$ denotes the optimal language-pair-relevant layers.


\subsection{Evaluating the Language Awareness of Neurons}
\label{Special Neurons Evaluation}

Once we find the language-pair-relevant layer, do we need to finetune all neurons of the layer for the corresponding language pair? 
Our experiments show that this all-neuron-finetuning strategy is not as expected (see Section \ref{Ablation Study}).
The main reasons are two fold. 
First, if all parameters of the detected FFNs are updated for all language pairs, catastrophic forgetting problem still remains \cite{DBLP:conf/aaai/LiuYW21}. 
Second, there is no effective mechanism to overcome the parameter interference issue to preserve the language-general and the language-specific knowledge.

Partially inspired by the studies on the importance-based neuron finetuning for NMT \cite{DBLP:conf/acl/Xie0G020} and neuron interpretability in LLMs \cite{DBLP:journals/corr/abs-2309-04827}, we propose to use the TE to evaluate which neurons are essential to all languages and which neurons are responsible for specific languages.
We first define the awareness score $\Phi(i)$ of a neuron to a certain language:

\begin{equation}
\Phi(i) = \left | \Delta \mathcal{L}(\textbf{h}_{i}) \right | , i \in {L_{j}}
\label{eq1}
\end{equation}
$L_{j}$ is the $j$-th layer that is the detected language-pair-relevant layer.
$\textbf{h}_{i}$ is the output of neuron $i$.
$\Delta \mathcal{L}(\textbf{h}_{i})$ is the loss change between setting $\textbf{h}_{i}$ to \textbf{0} and keeping it at its original value. 
It can be transformed by TE into the following form:

\begin{equation}
\left|\Delta \mathcal{L}\left(\textbf{h}_{i}\right)\right|=\left|\frac{\partial \mathcal{L}}{\partial \textbf{h}_{i}} \textbf{h}_{i}\right|
\label{eq2}
\end{equation}

We estimate the loss change as the product of the gradient of the loss function with respect to the activation value and the activation value itself. 
The detailed proof can be found in the appendix \ref{Taylor Expansion}, which is similar to that by \citet{DBLP:conf/acl/Xie0G020}.
Then, we determine which neurons are shared across all language pairs (i.e., language-general neurons) and which neurons are only related to specific language pairs.

We define $\mathbf{X}_i$ as the vector of awareness scores of the $i$-th neuron for each language. For each neuron, we calculate the variance $\sigma(\mathbf{X}_i)$ of the awareness scores across different languages. 
Within a specific layer, we sort the neuron awareness scores based on their variance from the highest to the lowest. 
A variance threshold $\lambda (i)$ is calculated to distinguish language-general neurons from language-specific neurons as follows:

\begin{equation}
\lambda (i) = \mathrm{sort} (\sigma(\mathbf{X_i}))_{\lfloor \epsilon \times p \rfloor}, i \in {L_{j}}
\label{eq3}
\end{equation}
where $p$ is the number of neurons in the $L_j$ layer, $\epsilon$ is a predefined ratio. 
For neurons with language awareness score variances below the estimated threshold $\lambda(i)$, we categorize them as language-general neurons, otherwise as language-specific neurons. 
Each detected language-specific neuron is assigned to the language with the highest awareness score.

The set of neurons that are specific either to the source language or to the target language are aggregated as the neurons exclusive to that language pair. 

\subsection{Routing and Finetuning}
\label{Routing and Finetuning}

\begin{figure}[t]
\centering
\centerline{\includegraphics[scale=0.5]{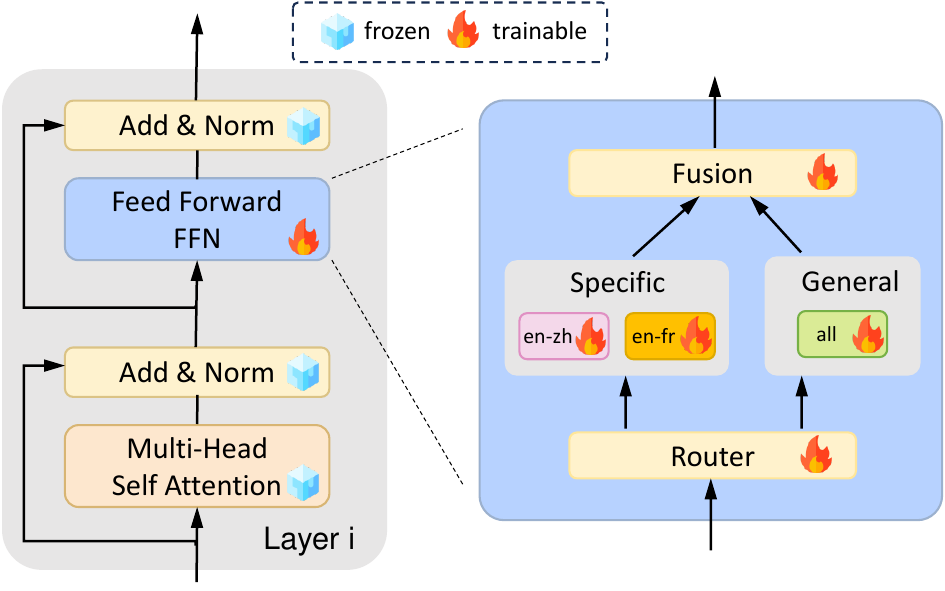}}
\caption{The model architecture used for routing and training.}
\label{fig2}
\end{figure}

In our proposed framework, for a given bilingual dataset of a specific language pair, only the language-general and language-specific neurons of the detected FFNs for this language pair participate in the forward computation and the parameters associated with them are updated during the backward propagation, as illustrated in Figure \ref{fig2}.
Nevertheless, it has been empirically shown that the language signals from language indicator tokens alone are not sufficient \cite{DBLP:journals/corr/abs-1907-05019}, making modules or mechanisms dedicated to language-general and language-specific modeling a necessity \cite{DBLP:conf/acl/ZhangWTS20,DBLP:conf/iclr/ZhangBSF21}.
To address this issue, we propose a conditional awareness-based routing mechanism (CAR) that allows the model to decide and learn what proportion of the outputs of language-general and language-specific neurons should be allocated for the translation of the language pair.
For an input token $x_{t}$, CAR is evaluated as follows:

\begin{equation}
\text{CAR}(x_{t}) = \frac{\sum_{i=1}^{N} \Phi (i)}{\sum_{i=1}^{N} \Phi (i)+\sum_{j=1}^{M} \Phi (j)}
\label{eq3}
\end{equation}

\begin{equation}
\textbf{h}_{G}(x_{t}) = \text{FFN}_{G}(\text{CAR}(x_{t}).x_{t})
\label{eq4}
\end{equation}

\begin{equation}
\textbf{h}_{S}(x_{t}) = \text{FFN}_{S}((1-\text{CAR}(x_{t})).x_{t})
\label{eq5}
\end{equation}
where $G$ denotes language-general and $S$ language-specific.
$\text{FFN}_{G}$ and $\text{FFN}_{S}$ are language-general and language-specific neurons, respectively.
$N$ is the total number of language-specific neurons in a FFN layers for a language pair.
$M$ is the total number of language-general neurons in a FFN layers.
We combine $\text{FFN}_{G}$ and $\text{FFN}_{S}$ to alleviate the parameter interference. 
The fusion output $\textbf{H}^{f}$ is given by:

\begin{equation}
\textbf{H}^{f} = \textbf{h}_{G}(x_{t}) + \textbf{h}_{S}(x_{t})
\label{eq6}
\end{equation}
Uppercase $\textbf{H}^{f}$ is just a notation here for the addition result of $\textbf{h}_{G}(x_{t})$ and $\textbf{h}_{S}(x_{t})$, which is only used to distinguish it from $\textbf{h}_{G}(x_{t})$ and $\textbf{h}_{S}(x_{t})$. 
During the finetuning stage, we only update the parameters of language-general and language-specific neurons for a specific language pair and freeze other parameters of LLMs.

\section{Experiments}
We conducted extensive experiments with involving multiple models across various translation directions to evaluate the proposed framework against a set of strong baselines. 

\subsection{Dataset}
During the finetuning stage, we selected 5 language pairs to tune LLMs. 
All the original training data came from the recent WMT general translation
track. 
All data followed the license that can be freely used for research purposes. 
In addition, we used the way in \cite{DBLP:conf/acl/HuangLMYL23} to clean training data. 
All datasets originated from the Workshop on Machine Translation (WMT)\footnote{https://www.statmt.org/}.
Specifically, we extracted 200,000 sentence pairs for each translation direction. In addition, we employed ten translation instruction finetuning templates sourced from FLAN v2 \cite{DBLP:conf/icml/LongpreHVWCTZLZ23}, adopting them to our parallel data. Each sentence pair from parallel corpus was randomly assigned one translation instruction template.
We assessed our model's performance using established test sets like WMT16, WMT14 and OPUS-100. 

\begin{table*}[t]
\centering
\resizebox{0.95\textwidth}{!}{%
\begin{tabular}{lllllllllllll}
\specialrule{0.1em}{1pt}{1pt}
\multicolumn{1}{l|}{\multirow{2}{*}{\textbf{Methods}}} & \multicolumn{1}{l|}{\multirow{2}{*}{\textbf{Params}}} & \multicolumn{2}{c|}{WMT16}         & \multicolumn{2}{c|}{WMT16}         & \multicolumn{2}{c|}{WMT14}         & \multicolumn{2}{c|}{OPUS-100}      & \multicolumn{2}{c}{OPUS-100} \\
\multicolumn{1}{l|}{} & \multicolumn{1}{l|}{} & en-de & \multicolumn{1}{l|}{de-en} & en-it & \multicolumn{1}{l|}{it-en} & en-fr & \multicolumn{1}{l|}{fr-en} & en-ar & \multicolumn{1}{l|}{ar-en} & en-zh & zh-en &     \\ \hline
\multicolumn{1}{l|}{Full finetuning}   &  \multicolumn{1}{l|}{7B}       & 16.12     & \multicolumn{1}{l|}{19.39}      &  17.98     & \multicolumn{1}{l|}{24.18}      &   29.13    & \multicolumn{1}{l|}{28.15}      &   15.40    & \multicolumn{1}{l|}{28.83}      &  20.87             &     25.52  \\
\multicolumn{1}{l|}{0-shot}   &  \multicolumn{1}{l|}{——}       & 11.71  & \multicolumn{1}{l|}{17.82}      & 8.07  & \multicolumn{1}{l|}{16.05}      & 19.58 & \multicolumn{1}{l|}{18.88}      & 9.46 & \multicolumn{1}{l|}{26.37}      & 6.14         & 22.02              \\
\multicolumn{1}{l|}{In-context} &   \multicolumn{1}{l|}{——}    & 11.49  & \multicolumn{1}{l|}{14.57}      & 9.80  & \multicolumn{1}{l|}{13.12}      & 18.74 & \multicolumn{1}{l|}{15.29}      & 10.01  & \multicolumn{1}{l|}{21.24}      & 6.83         & 17.19              \\
\multicolumn{1}{l|}{Adapter}   &     \multicolumn{1}{l|}{806M}         &  15.61     & \multicolumn{1}{l|}{19.67}      &  15.25     & \multicolumn{1}{l|}{23.63}      & 28.08      & \multicolumn{1}{l|}{27.26}      &   11.28    & \multicolumn{1}{l|}{28.07}      &  15.18             &     25.05         \\
\multicolumn{1}{l|}{LoRA}    &     \multicolumn{1}{l|}{31M}          &  15.12     & \multicolumn{1}{l|}{19.03}      &   14.82    & \multicolumn{1}{l|}{22.85}      &  27.16     & \multicolumn{1}{l|}{27.72}      &    11.15   & \multicolumn{1}{l|}{27.74}      &   15.72            &     25.12         \\
\multicolumn{1}{l|}{Adapter-LoRA}  &   \multicolumn{1}{l|}{806M}              &  16.31     & \multicolumn{1}{l|}{20.23}      &   15.83    & \multicolumn{1}{l|}{23.82}      &  28.05     & \multicolumn{1}{l|}{28.22}      &    11.78   & \multicolumn{1}{l|}{28.46}      &   16.32            &     25.61         \\
\multicolumn{1}{l|}{LANDeRMT (Ours)}   &    \multicolumn{1}{l|}{805M}          &  \textbf{18.85 }    & \multicolumn{1}{l|}{\textbf{22.03}}      &   \textbf{19.82}    & \multicolumn{1}{l|}{\textbf{25.99}}      &  \textbf{31.91}     & \multicolumn{1}{l|}{\textbf{30.55}}      &  \textbf{16.97 }    & \multicolumn{1}{l|}{\textbf{31.44}}      &  \textbf{22.47}             &  \textbf{28.11}            \\  
\specialrule{0.1em}{1pt}{1pt}
\end{tabular}}
\caption{BLEU scores on the 10 language pairs for xx-to-English and English-to-xx translation.
The highest score on each translation direction is highlighted in bold.}
\label{tab:table1}
\end{table*}

\subsection{Settings and Baselines}

\paragraph{Settings} 

In the language-pair-relevant layers detection phase, we set $k$ to 4. 
We executed a freezing operation on the parameters of the remaining layers while exclusively finetuning the parameters within the chosen four layers. 
In the language-aware neurons evaluation phase, we categorized the parameters within the selected layer into language-general and language-specific parameters, setting $\epsilon$ to 0.9. 
During the model finetuning stage, we configured the fintuning hytper-parameters as follows: the finetuning epoch was set to 1, the number of language pairs was specified as 10, the number of iterations per epoch for each language pair was set to 12,500, the batch size was set 8, and the AdamW optimizer was employed. 
Additionally, the learning rate was set to 5e-5. 
Furthermore, we introduced a gradient accumulation operation, updating the model parameters every 10 iterations to enhance convergence. 
The LLMs used for our experiments are BLOOM-7b1 \cite{DBLP:journals/corr/abs-2211-05100} and Baichuan2-7B-Base \cite{DBLP:journals/corr/abs-2309-10305}.

\paragraph{Baselines} We compared our approach to the following strong baselines.
\begin{itemize}
\item 0-shot: This approach uses instructions directly to make the model perform downstream tasks without providing any in-context demonstrations.
\item In-context \cite{DBLP:conf/icml/0006HB23}: This is a training-free approach that allows the LLMs to perform downstream tasks. In particular, we use 5 random shots as in-context demonstrations.
\item Adapter: This method facilitates the acquisition of new knowledge by incorporating additional adapter modules following model-specific layers, effectively addressing the issue of catastrophic forgetting.
\item LoRA \cite{DBLP:conf/iclr/HuSWALWWC22}: This method efficiently finetunes a model for a downstream task by converting certain structures into low-rank matrices and subsequently finetuning them to suit the task. 

\item Adapter-LoRA \cite{DBLP:conf/emnlp/AlvesGAPRSCM23}: It uses adapter-based finetuning with LoRA, which matches the performance of traditional finetuning while reducing the number of training parameters. 
\end{itemize}



\subsection{Main Results}
\label{Main Results}

For evaluating translation performance, we used two automatic evaluation metrics sacreBLEU\footnote{BLEU+case.mixed+numrefs.1+smooth.none+tok.13a\\+version.2.2.1}.

\textbf{Comparison with ICL}
In order to examine the effectiveness of our proposed method, we evaluated LANDeRMT on various test set and multiple language pairs.
As shown in Table \ref{tab:table1}, our method can use new parallel training data to enhance the translation ability of LLMs.


\textbf{Comparison with finetuning baselines}
Compared to the baselines, our method is the best for all translation directions in Table \ref{tab:table1}. 
For relatively low-resource language pairs, such as English-Chinese, our method achieves significant improvements over baselines. 
Compared to the full parameter finetuning approach, our method has a clear parametric advantage since it only fine-tunes parameters in four layers in the model.
Our approach exhibits a notable advantage in terms of the number of parameters to be tuned. 
In comparison to other efficient finetuning methods, e.g., the adapter baseline approach, our method finetunes basically the same number of parameters as it. 
However, our method is much better than the adapter-based approach in terms of translation quality.

\subsection{Ablation Study}
\label{Ablation Study}

\begin{table}[]
\centering
\resizebox{0.45\textwidth}{!}{%
\begin{tabular}{ccccc}
\toprule
           & en-fr & fr-en  & en-zh & zh-en \\ \midrule
Layers     & 27.63      & 27.12       & 22.81      & 24.28      \\
LANDeRMT-LS    &   22.27    &  21.46      &  16.18     &  23.16     \\
LANDeRMT-LG    &    28.15   &   27.83     &  23.52     & 25.08      \\
LANDeRMT (Ours) &  31.91     &  30.55      & 22.47      & 28.11       \\ \bottomrule
\end{tabular}%
}
\caption{Translation results achieved by finetuning the BLOOM-7b1 model using different ablation experiment settings.}
\label{tab:table2}
\end{table}

In the ablation experiments, we employed four distinct experimental setups, denoted as Layers, LANDeRMT-LS, LANDeRMT-LG, and LANDeRMT. 
The Layers configuration finetuned all parameters of the lanuage-pair-relevant layer for each language direction. 
In the LANDeRMT-LS setup, we finetuned only the language-specific parameters of the selected layers, with each language direction adjusting parameters specific to that language direction.
The LANDeRMT-LG setup focused on finetuning only the language-general parameters of the selected layers, with all language directions adjusting the same language-general parameters.
The LANDeRMT method, proposed in this paper, finetuned both the language-general parameters and language-sepecific parameters to each language pair.

From Table \ref{tab:table2}, we observe that LANDeRMT-LS underperforms the Layers method, likely due to its smaller parameter size, which constitutes only 10\% of the parameters in Layers. 
In details, we can observe that LANDeRMT achieves a 4.28 BLEU improvement over Layers in en-fr, a 9.64 BLEU improvement over LANDeRMT-LS, and a 3.76 BLEU improvement over LANDeRMT-LG. 
These experiments demonstrate the effectiveness of CAR.
Surprisingly, LANDeRMT-LG achieves better results despite finetuning fewer parameters than Layers. This suggests that our selection of language-general parameters effectively captures language alignment, significantly improving translation performance.
However, finetuning the language-general parameters alone, as in LANDeRMT-LG, is insufficient to fully grasp language-specific information. 

\begin{figure}[t]
\centering
\centerline{\includegraphics[scale=0.5]{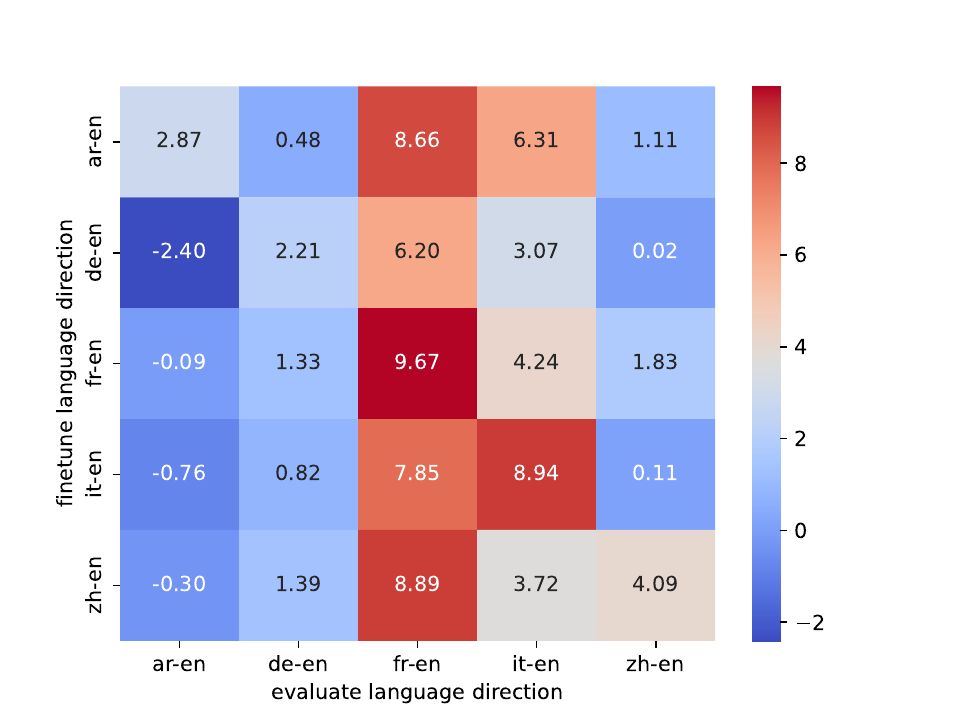}}
\caption{BLEU scores improvement achieved on other language pairs using the LANDeRMT method for finetuning only one language pair on the BLOOM-7b model.}
\label{fig3}
\end{figure}

\section{Analysis}

\subsection{LANDeRMT Improves Transfer Learning across Languages}
\label{EINA Improves Language Alignment and transfer Learning}

We examined the transfer learning ability of LANDeRMT in different translation directions. 
We finetuned the model using only parallel data from a particular language direction. In other words, we finetuned only the language-general and language-specific parameters for that language pair, and then observed the performance of the model in other language directions. 
The Y-axis of Figure \ref{fig3} shows the single language direction that we have finetuned, and the X-axis shows the language direction of the test data, which is plotted as the improvement in the model's translation performance before and after the finetuning. 
Since BLOOM-7b is a model that is not mainly trained on a parallel corpus, its translation performance before finetuning is poor, which is the reason for the large improvement in the model's translation performance. 
We observe that when finetuning one language direction, the results of other language directions can also be significantly improved, which proves that our method is effective in facilitating transfer learning between languages. 
However, there are some exceptions. 
For example, when finetuning the de-en direction, the ar-en direction does not improve significantly or even decrease to some extent. 
We believe that this may be due to the fact that Arabic belongs to a different language family from German, and that the distance between the languages is far, making it difficult for cross-lingual transfer learning.

\begin{figure}[t]
\centering
\centerline{\includegraphics[scale=0.38]{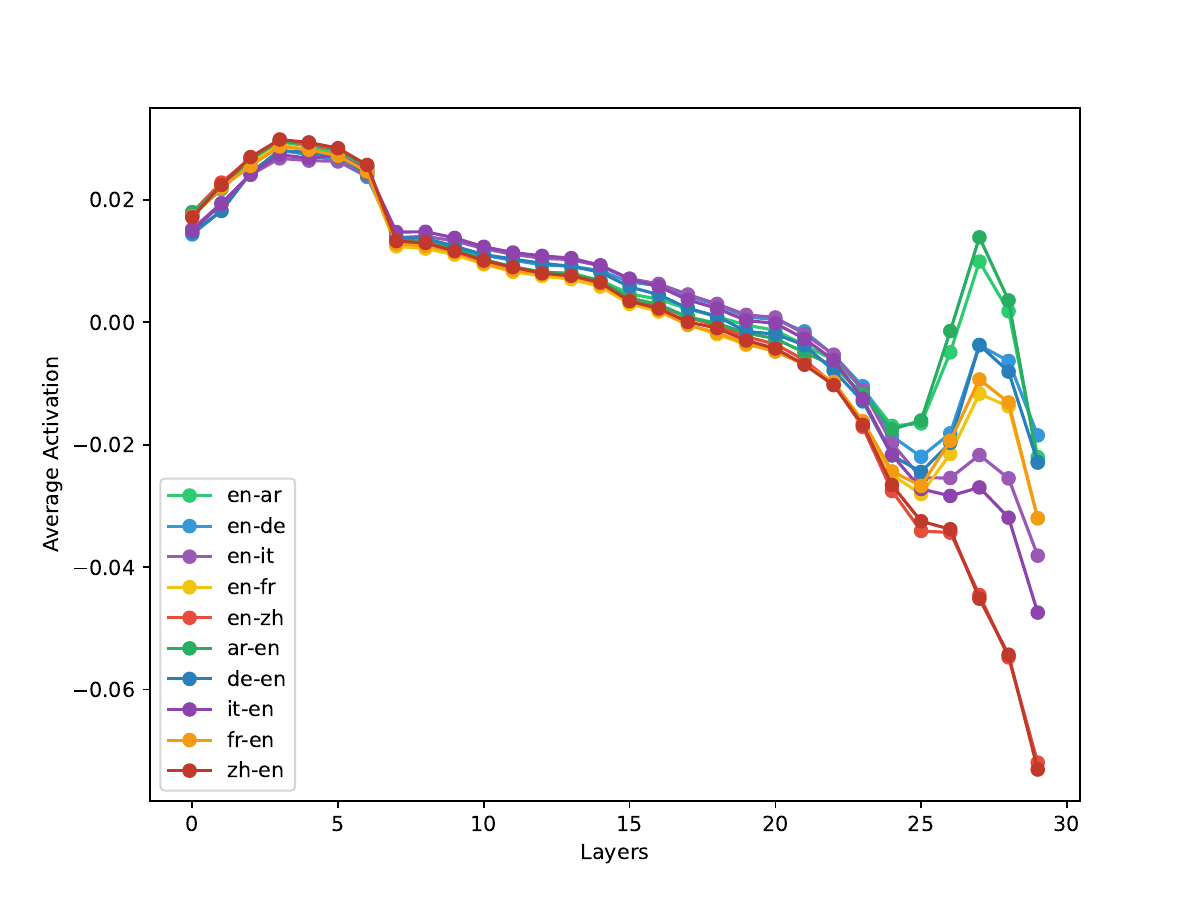}}
\caption{Layer-wise average activation across various language pair settings in the BLOOM-7b model.}
\label{fig4}
\end{figure}

\begin{figure}[t]
\centering
\centerline{\includegraphics[scale=0.38]{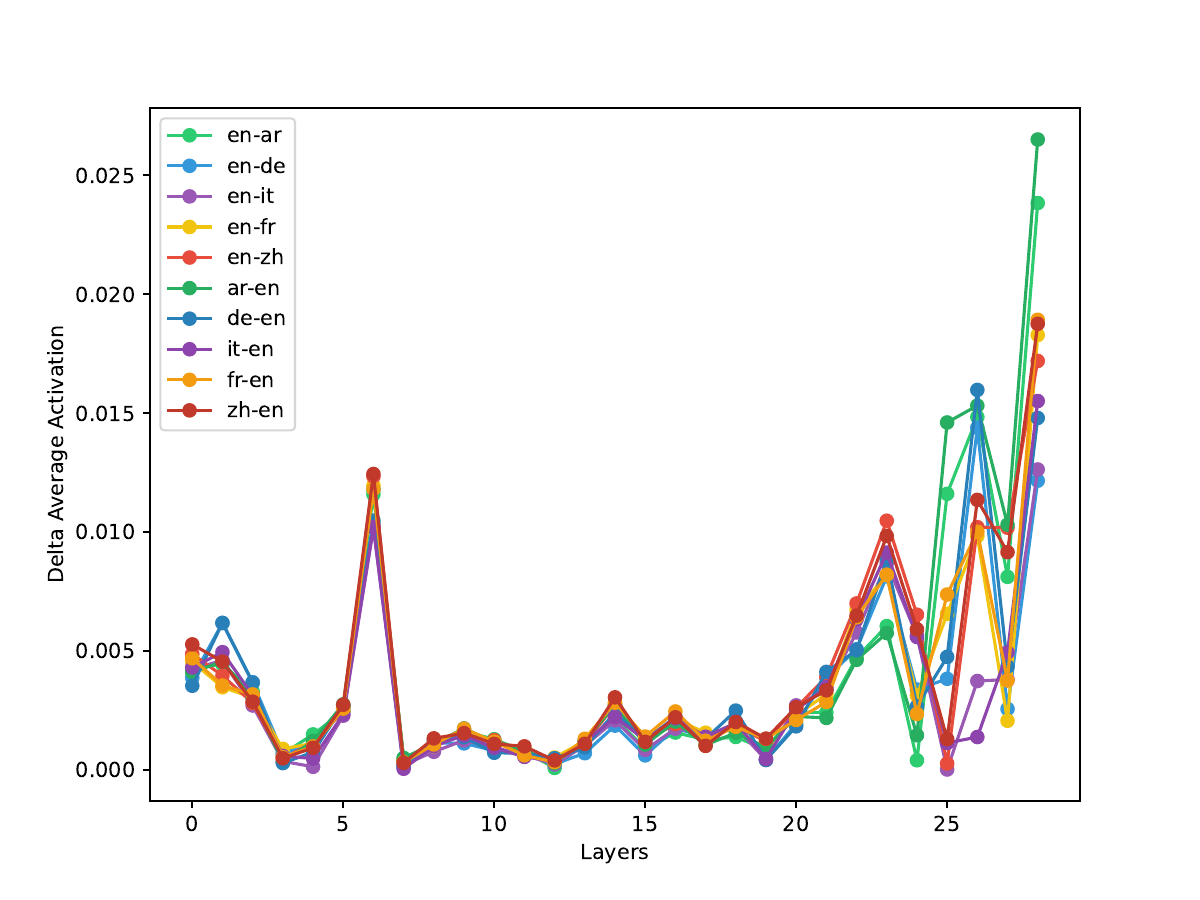}}
\caption{Layer-wise delta average activation across various language pair settings in the BLOOM-7b model.}
\label{fig5}
\end{figure}

\subsection{Language-Pair-Relevant Layers for Different Language Pairs}

\begin{figure*}[t]
\centering
\centerline{\includegraphics[scale=0.41]{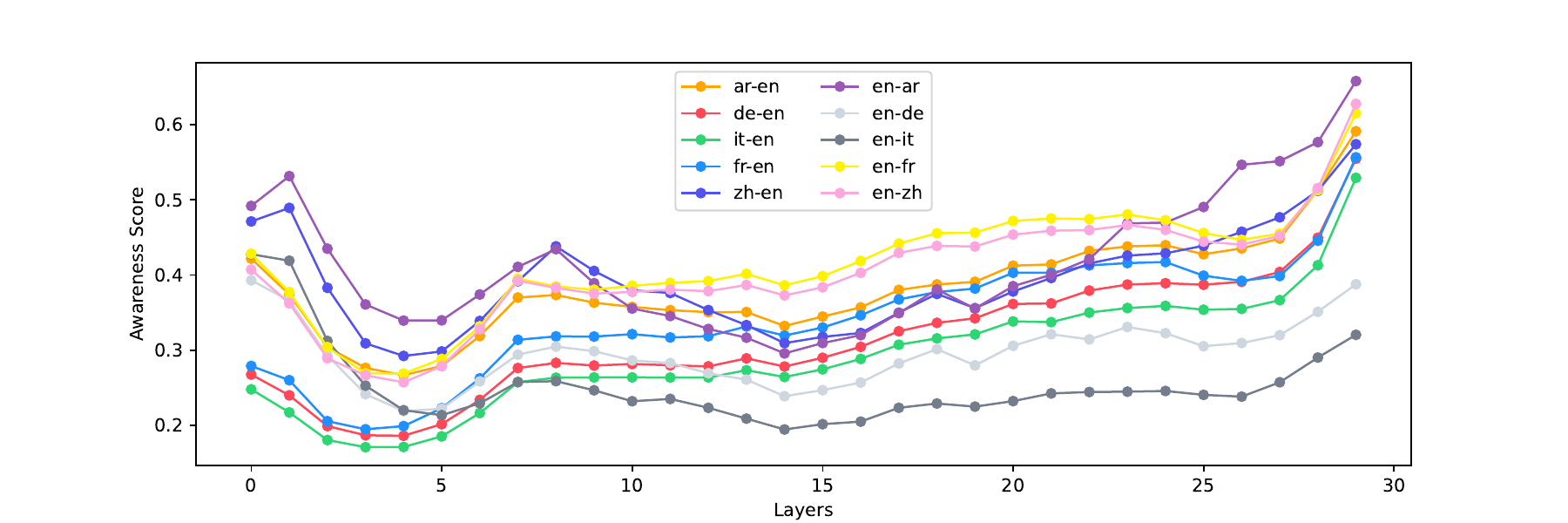}}
\caption{Layer-wise average language-general neuron awareness scores across various language settings in the BLOOM-7b1 model.}
\label{appd:fig1}
\end{figure*}

\begin{figure*}[t]
\centering
\centerline{\includegraphics[scale=0.41]{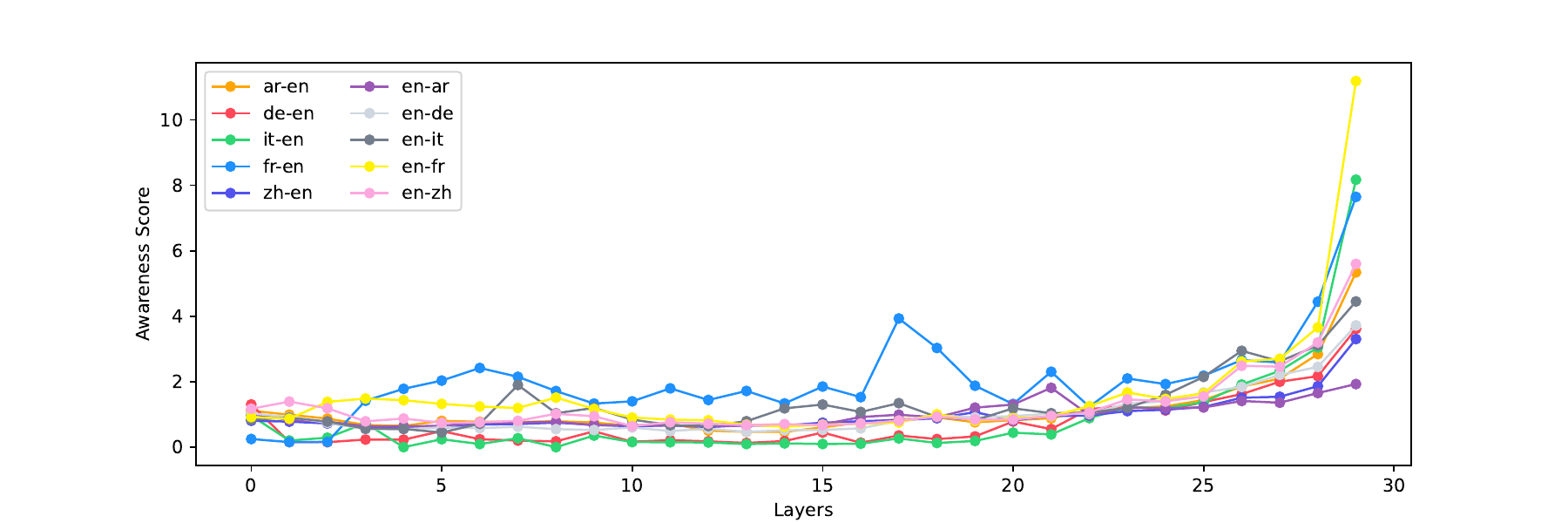}}
\caption{Layer-wise average language-specific neuron awareness scores across various language settings in the BLOOM-7b1 model.}
\label{appd:fig2}
\end{figure*}

Figure \ref{fig4} illustrates the variation in the average activation values across each layer of the model when inputting translation instructions generated using diverse language pairs.  
It is noteworthy that the average activation values of various language pairs exhibit a similar trend of change, particularly in the shallower layers of the model. 
Moreover, when interchanging the source and target languages within language pairs, the average activation values consistently follow a more uniform trend, as evidenced by the translation directions of ar-en and en-ar.
This indicate that not only the semantic remains consistent between the source and target languages within the same language pair, but the identical semantic is still existing on across different language pairs.

The absolute value of activation value changes from layer to layer can be calculated by using the average activation values of each layer, as illustrated in Figure \ref{fig5}. We can find that early and late-stage layers in the model harbor information pertaining to language pairs. 
For instance, layers 6 and 7, as well as the final layers, exhibit higher absolute values of activation value changes. 
Furthermore, an observation can be made that the early layers of the model encapsulate language pair-related information that is language pair-general, displaying a substantial overlap across different language pairs.
Conversely, the layers towards the end of the model contain language pair-related information that is language pair-specific, characterized by a diminished overlap among different language pairs.

\begin{figure*}[t]
\centering
\centerline{\includegraphics[scale=0.38]{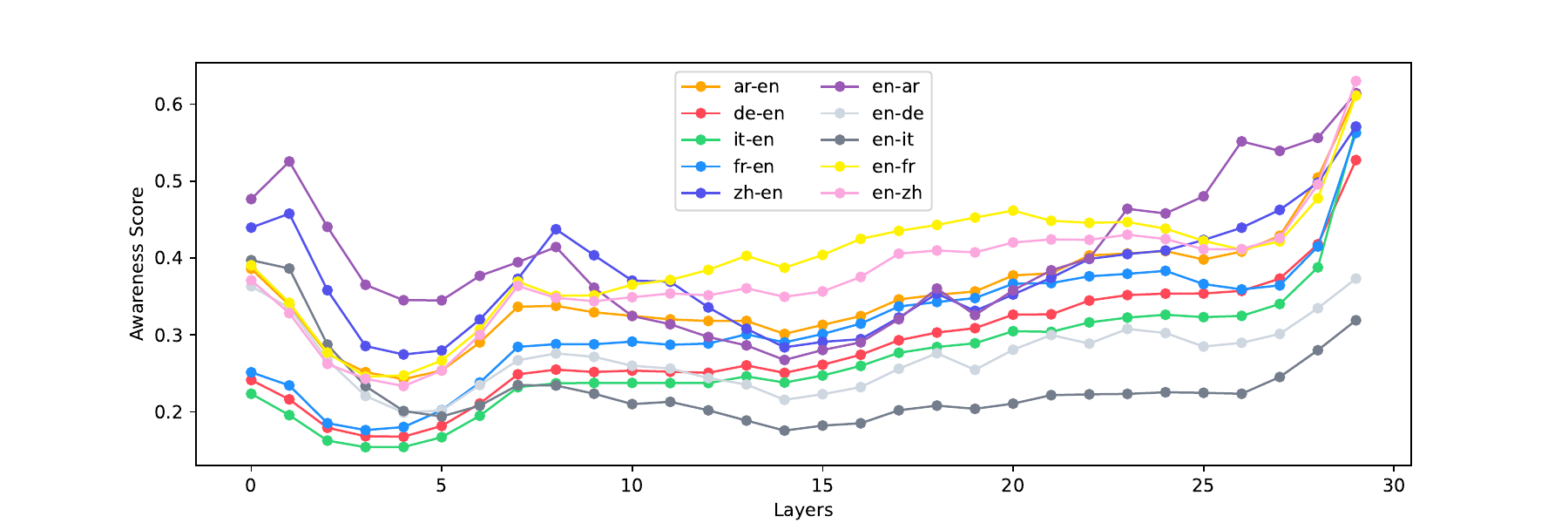}}
\caption{Layer-wise average neuron awareness scores across various language settings in the BLOOM-7b model.}
\label{fig6}
\end{figure*}

\subsection{Neuron Awareness for Different Languages: General and Specific}
\label{Neuron Importance for Different Languages: General and Specific}

The main idea of our proposed method is to distinguish language-general from language-specific neurons.
To verify whether this goal has been achieved, we conducted the following experiments. 
As mentioned in Section \ref{Special Neurons Evaluation}, we categorize neurons based on their awareness scores, and we observed significant differences in the awareness scores of language-general and language-specific neurons across layers requiring fine-tuning. 
We illustrate this in Figure \ref{appd:fig1} and Figure \ref{appd:fig2}.
We can find that for almost all language pairs, there are noticeable differences in awareness scores between certain intermediate layers and the final layers. 
This indicates that our categorization of neurons based on layers accurately reflects the practical scenario.
It also suggests that language-general and language-specific neurons exhibit varying levels of importance across different layers of the model, particularly in layers targeted for finetuning. 
Such differences likely stem from the distinct roles that language-general and language-specific neurons play in capturing and processing language-specific and language-general information.

\subsection{Neuron Awareness for Different Language Pairs}
\label{Neuron Importance for Different language pairs}



Figure \ref{fig6} depicts the average neuron awareness scores for each layer of the model, computed using TE with monolingual data inputs in various language pairs. 
The employed multidirectionally aligned monolingual data ensures semantic one-to-one correspondence. 
The results show consistent trends in neuron awareness scores across different language pairs, particularly in the intermediate layers, indicating the model's ability to capture semantic information consistently across language pairs. 
Additionally, related languages such as Spanish, English, and French exhibit more similar trends, supporting our hypothesis.

Furthermore, we observed that language-specific neurons tend to have higher awareness scores in the last layers of the model. 
This suggests a heightened focus on encoding and retaining language-specific semantic information during the output phase, particularly in deeper layers.
Notably, language-specific neurons related to English consistently exhibit high awareness scores across all language pairs. 
This can be attributed to the prevalence of English data during the model's pre-training phase, indicating robust representation and preservation of English language-specific information throughout the model.

\begin{figure}[t]
\centering
\centerline{\includegraphics[scale=0.46]{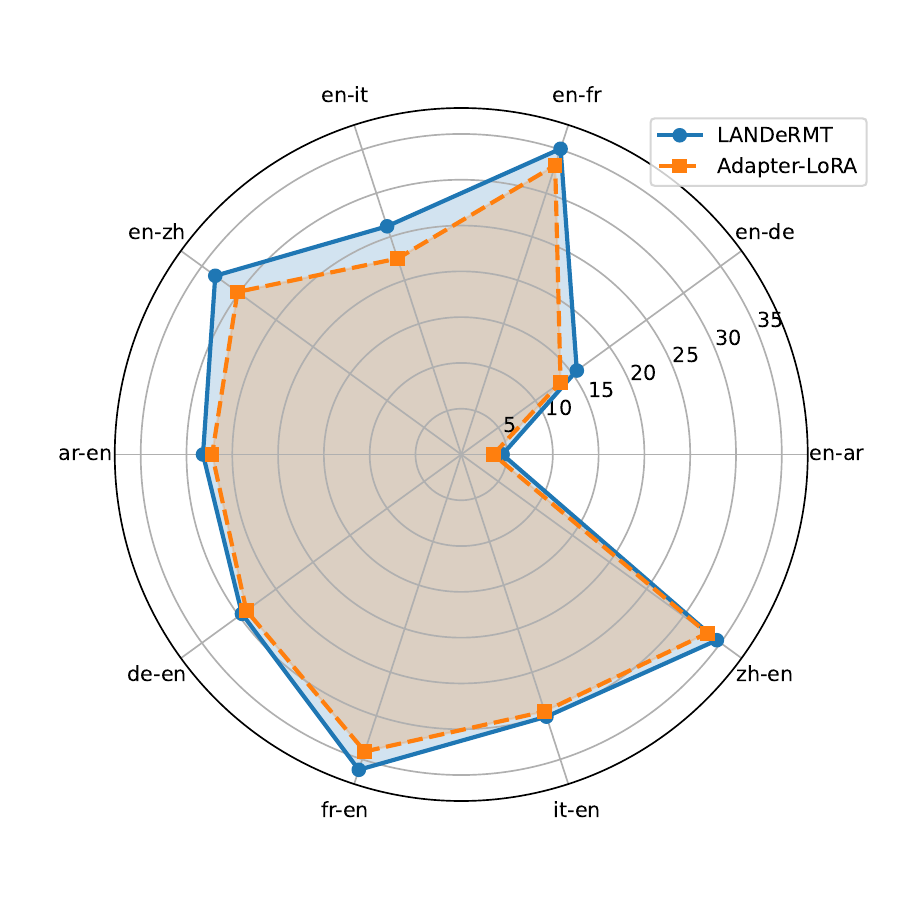}}
\caption{Comparison of BLEU scores on the OPUS 100 test set across ten language directions for finetuning the Baichuan-7b-base model using adapter-LoRA and our proposed method LANDeRMT.}
\label{fig8}
\end{figure}

\subsection{Results on Other LLMs}
\label{Results on Different LLMs}


We also finetuned the Baichuan-7B-Base model using the LANDeRMT method and compared it with the adapter-LoRA finetuning approach. 
Results are shown in Figure \ref{fig8}.
We observe that across the 10 language directions selected our proposed method outperforms the adapter-LoRA finetuning method. 
This demonstrates the applicability of the LANDeRMT method across different models, achieving optimal results not only in the BLOOM-7b models but also in the Baichuan model.


\subsection{Effect of Hyperparameter $k$}

\begin{figure}[t]
\centering
\centerline{\includegraphics[scale=0.35]{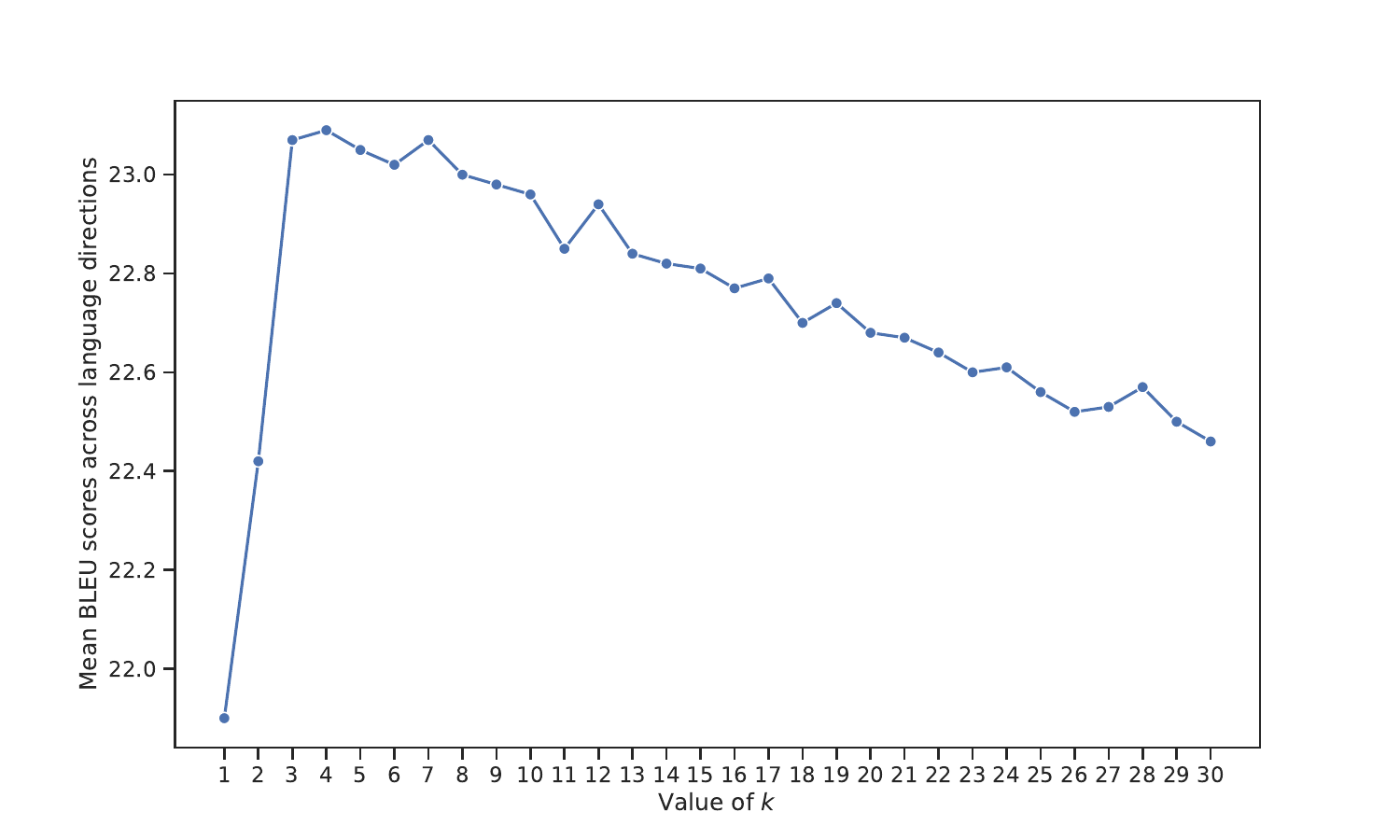}}
\caption{Mean BLEU scores for all language directions at different $k$ value settings.}
\label{fig-k}
\end{figure}

When the relation of layers for different languages is determined, the number of language pairs associated with each layer can be adjusted according to k.
When $k$ = 30, the threshold is max, so all layers will be allocated to tune LLMs, and when k = 0, the threshold is 0 so none layers will be tuning for all language pairs just like the 0-shot ICL.
To better show the overall impact of the hyperparameter $k$, we vary it from 0 to 30 and the results are shown in Figure \ref{fig-k}.
As we can see, the translation performance of the proposed approach increases with the increment of $k$ and reach the best performance when $k$ equals 4. 
As $k$ continues to increase, the performance deteriorates, which indicates that the over-specific layers are bad at capturing the common language-pair-relevant alignment and will lead to performance degradation.

\subsection{Effect of Hyperparameter $\epsilon$}

\begin{figure}[t]
\centering
\centerline{\includegraphics[scale=0.5]{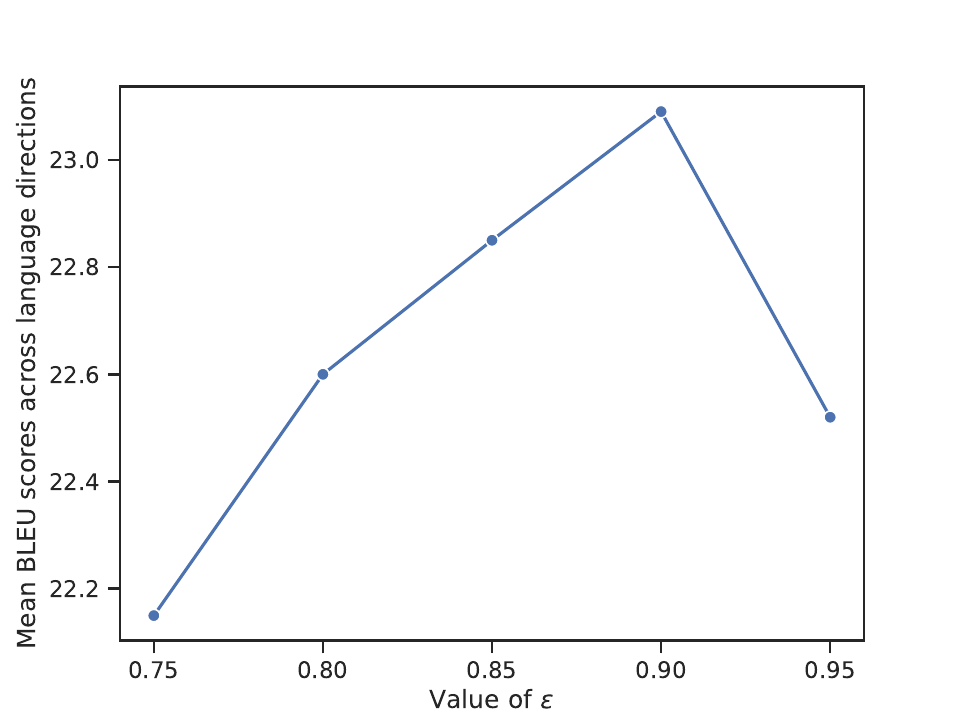}}
\caption{Mean BLEU scores for all language directions at different $\epsilon$ value settings.}
\label{fig7}
\end{figure}

We set several different sets of $\epsilon$ values to classify language-general neurons and language-specific neurons. The experimental outcomes are depicted in Figure \ref{fig7}. The figure reveals that the average value of all language-specific BLEU scores peaks when language-general neurons constitute 0.9 of the total neuron count. Below this threshold, the translation efficacy diminishes as the proportion of language-general neurons decreases. Conversely, exceeding the 0.9 threshold results in a decline in performance, with a higher proportion of language-general neurons leading to poorer results.

\begin{figure}[t]
\centering
\centerline{\includegraphics[scale=0.4]{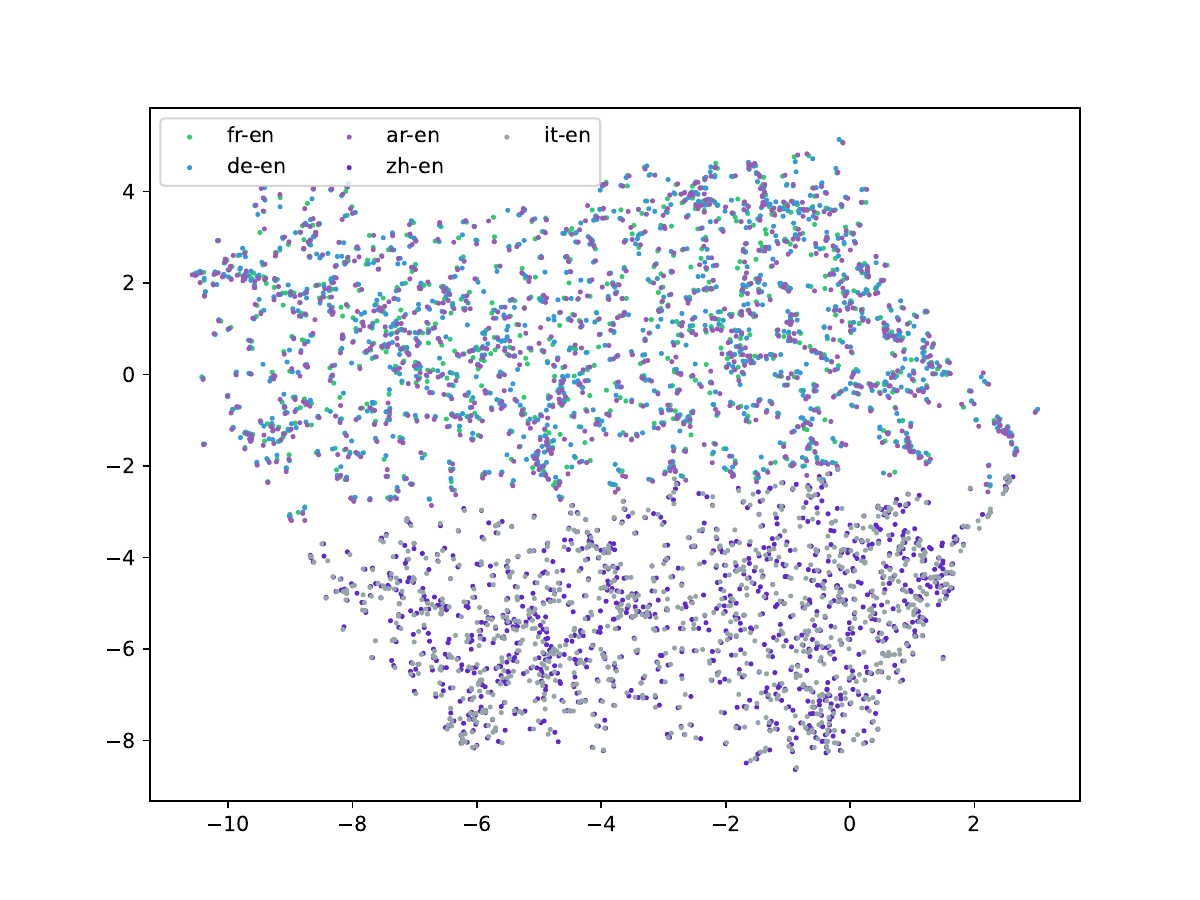}}
\caption{Clustering of representations generated by language-general neurons in the mlp.dense\_h\_to\_4h structure in layer 10 of the BLOOM-7b1 model.}
\label{appd:fig3}
\end{figure}

\begin{figure}[t]
\centering
\centerline{\includegraphics[scale=0.4]{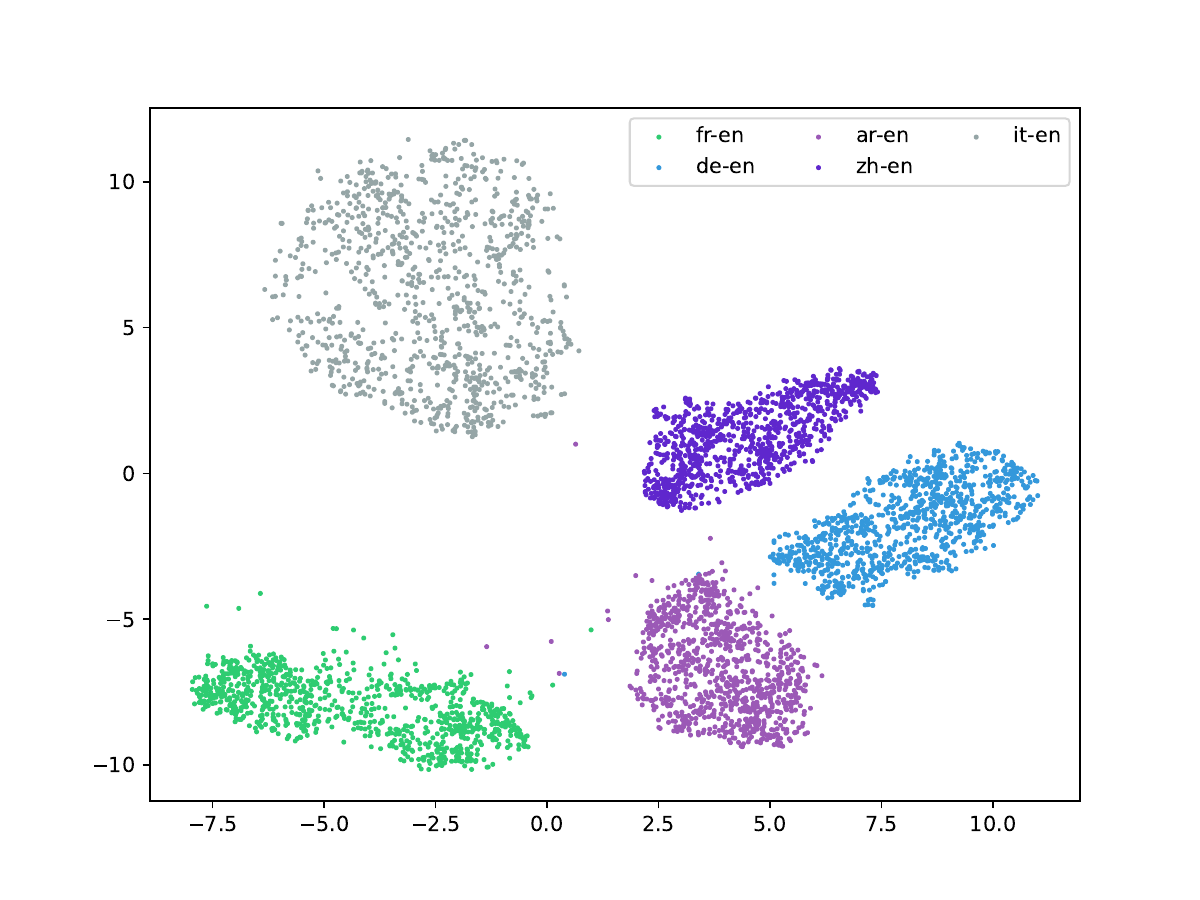}}
\caption{Clustering of representations generated by language-specific neurons in the mlp.dense\_h\_to\_4h structure in layer 10 of the BLOOM-7b1 model.}
\label{appd:fig4}
\end{figure}

\subsection{Language Cluster}

The main idea of our proposed method is to let the language-general and the language-specific knowledge be captured by different neurons.
To validate whether the language-general and language-specific neurons of FFNs within LLMs general or specific language knowledge, we plotted the distribution of different neurons across various languages in $10$-th layer, as shown in Figure \ref{appd:fig3} and Figure \ref{appd:fig4}. 
From these figures, it is evident that for the language-general FFNs neurons, the distributions for various languages intersect without clear boundaries, indicating a shared representation of language knowledge. 
In contrast, for the language-specific neurons, the boundaries between language distributions are highly distinct, highlighting the independence of language-specific knowledge.
This observation underscores our neurons awareness can make neurons capture and integrate language knowledge in a general manner across multiple languages within the language-general FFNs. 
Conversely, the distinct boundaries in the distributions of language-specific FFNs neurons suggest that these neurons are dedicated to encoding language-specific nuances and characteristics.

\section{Conclusion}

In this paper, we have presented a novel approach that not only improves translation quality but also mitigates the risk of forgetting previous knowledge while adapting to new data.
We propose a TE to evaluate neuron awareness scores for MT tasks and categorize them into language-general neurons and language-specific neurons.
The proposed routing mechanism ensures optimal allocation of resources across language-specific and language-general capacities, further enhancing the adaptability of LLMs.
Our experimental results, conducted across ten language pairs, validate the effectiveness of our model, showcasing superior performance compared to existing baselines.

\section*{Acknowledgments}

The present research was supported by the National Natural Science Foundation of China Youth Foud (Grant No.62306210) and the Key Research and Development Program of Yunnan Province (Grant No. 202203AA080004). 
We would like to thank the anonymous reviewers for their insightful comments.

\section*{Limitations}
Although LANDeRMT is a new approach to finetune LLMs to enhance the translation ability of LLMs. 
The finetuning procedure is shorter than training LLMs as the amount of data required during the finetuning stage is much smaller than during the training stage. 
This significantly reduces the cost of training model from scratch but maybe still totally overcome parameters interference as we not fully update the parameters of LLMs.
Additionally, due to computational constraints, we are currently unable to design additional experiments to validate how our method enhances the upper limit of translation capabilities of LLMs when more training data is added.

\section*{Ethics Statement}
This study adheres to the ethical guidelines set forth by our institution and follows the principles outlined in the ACM Code of Ethics and Professional Conduct. 
All datasets used in our experiments are publicly available.

\bibliography{anthology}

\clearpage

\appendix

\section{Appendix}
\label{sec:appendix}

\subsection{Language Details}

We introduce the characteristics of different languages as shown in Table \ref{tab-lang}.

\begin{table}[ht]
\centering
    \begin{tabular}{cccc}
    \hline
    \textbf{Code} & \textbf{Language} & \textbf{Genus} & \textbf{Order} \\ \hline
    en            & English            & Romance        & SVO            \\
    ar            & Arabic            & Semitic        & VSO            \\
    fr            & French            & Romance        & SVO            \\
    de            & German            & Germanic       & SVO            \\
    zh            & Chinese           & Sinitic        & SVO            \\
    it            & Italian          & Romance        & SVO            \\
   \hline
    \end{tabular}
    \caption{The characteristics of languages in our setting.}
\label{tab-lang}
    \end{table}

\subsection{Taylor Expansion}
\label{Taylor Expansion}

We first express $\Delta \mathcal{L}(\textbf{h}_{i})$ as loss change as shown in the following equation.

$$
\left|\Delta \mathcal{L}\left(\textbf{h}_{i}\right)\right|=\left|\mathcal{L}\left(\textbf{H}, \textbf{h}_{i}=\textbf{0}\right)-\mathcal{L}\left(\textbf{H}, \textbf{h}_{i}\right)\right|
$$

$\textbf{H}$ is the representation produced by a neuron other than $i$ in the same structure as the $i$ neuron. We then perform a first-order Taylor expansion of $\mathcal{L}(\textbf{H},\textbf{h}_{i})$ at $\textbf{h}_{i}=\textbf{0}$.

$$
\mathcal{L}\left(\textbf{H}, \textbf{h}_{i}\right)=\mathcal{L}\left(\textbf{H}, \textbf{h}_{i}=\textbf{0}\right)+\frac{\partial \mathcal{L}\left(\textbf{H}, \textbf{h}_{i}\right)}{\partial \textbf{h}_{i}} \textbf{h}_{i}+R_{1}\left(\textbf{h}_{i}\right)
$$

The term $R_{1}\left(\textbf{h}_{i}\right)$ can be ignored since the derivatives of the activation function of second order and higher in the model tend to zero. So the above equation can be reduced to the following form.

$$
\mathcal{L}\left(\textbf{H}, \textbf{h}_{i}\right)\approx \mathcal{L}\left(\textbf{H}, \textbf{h}_{i}=\textbf{0}\right)+\frac{\partial \mathcal{L}\left(\textbf{H}, \textbf{h}_{i}\right)}{\partial \textbf{h}_{i}} \textbf{h}_{i}
$$

Therefore $|\Delta \mathcal{L}(\textbf{h}_{i})|$ can eventually be simplified to the following form.

$$
\left|\Delta \mathcal{L}\left(\textbf{h}_{i}\right)\right| \approx \left|\frac{\partial \mathcal{L}\left(\textbf{H}, \textbf{h}_{i}\right)}{\partial \textbf{h}_{i}} \textbf{h}_{i}\right|
$$

\end{document}